%% file: main.tex
\documentclass[10pt,twocolumn,letterpaper]{article}

\usepackage[]{cvpr}      
\usepackage{amsmath}
\usepackage{graphicx}
\usepackage{setspace}
\usepackage{multirow}
\usepackage{threeparttable}


\definecolor{cvprblue}{rgb}{0.21,0.49,0.74}
\usepackage[pagebackref,breaklinks,colorlinks,allcolors=cvprblue]{hyperref}

\def\paperID{*****} 
\def\confName{CVPR}
\def\confYear{2025}

\title{FVOS for MOSE Track of 4th PVUW Challenge: 3rd Place Solution}

\author{Mengjiao Wang, Junpei Zhang, Xu Liu, Yuting Yang, Mengru Ma\\ \\
International Joint Research Center for Intelligent Perception and Computation, Xi'an, China\\}

\begin{document}
\maketitle
\input{sec/0_abstract}    
\input{sec/1_intro}

\input{sec/2_formatting}

\input{sec/3_finalcopy}
\input{sec/4_conclusion}
{
    \small
    \bibliographystyle{ieeenat_fullname}
    \bibliography{main}
}


\end{document}

%% file: sec/0_abstract.tex
\begin{abstract}
Video Object Segmentation (VOS) is one of the most fundamental and challenging tasks in computer vision and has a wide range of applications. Most existing methods rely on spatiotemporal memory networks to extract frame-level features and have achieved promising results on commonly used datasets. However, these methods often struggle in more complex real-world scenarios. This paper addresses this issue, aiming to achieve accurate segmentation of video objects in challenging scenes. We propose fine-tuning VOS (FVOS), optimizing existing methods for specific datasets through tailored training. Additionally, we introduce a morphological post-processing strategy to address the issue of excessively large gaps between adjacent objects in single-model predictions. Finally, we apply a voting-based fusion method on multi-scale segmentation results to generate the final output. Our approach achieves $\mathcal{J\&F}$ scores of 76.81\% and 83.92\% during the validation and testing stages, respectively, securing third place overall in the MOSE Track of the 4th PVUW challenge 2025.
\end{abstract}

%% file: sec/1_intro.tex
\section{Introduction}
\label{sec:intro}

In recent years, with the development of deep learning, fundamental computer vision tasks such as image classification, object detection, and semantic segmentation have been effectively addressed through the introduction of numerous methods. However, an increasing amount of visual data is presented in the form of videos, where scenes enriched with both temporal and spatial information have garnered significant attention from researchers. In the domain of video-based visual analysis, Video Object Segmentation (VOS) stands out as one of the most fundamental yet challenging tasks. VOS aims to segment specific objects across an entire video sequence and plays a crucial role in many real-world applications that involve video analysis and understanding, such as autonomous driving, augmented reality, video editing, and more. Video object segmentation encompasses various settings, such as semi-supervised settings where the segmentation mask of the target object in the first frame is given, unsupervised settings where the primary object is automatically identified, and interactive settings that rely on user interactions to define the target object. This paper focuses on the most common semi-supervised VOS setting, and unless otherwise specified, VOS in this context refers to semi-supervised video object segmentation.

Existing methods for VOS have achieved high performance on the two most widely used public datasets, DAVIS \cite{pont20172017} and YouTube-VOS \cite{xu2018youtube}. For example, QDMN++ \cite{liu2025learning} achieves a $\mathcal{J\&F}$ score of 92.3\% on DAVIS 2016 and an overall $\mathcal{G}$ score of 86.6\% on YouTube-VOS 2018. Similarly, Cutie \cite{cheng2024putting} achieves a $\mathcal{J\&F}$ score of 88.8\% on DAVIS 2017 and an overall $\mathcal{G}$ score of 86.5\% on YouTube-VOS 2019. However, their performance on more challenging and complex datasets, such as MOSE \cite{ding2023mose}, is less satisfactory. The MOSE dataset features complex real-world scenarios involving object disappearance and reappearance, small or inconspicuous objects, severe occlusion, and crowded scenes. On this dataset, C-QDMN++ \cite{liu2025learning} achieves a $\mathcal{J\&F}$ score of 64.6\%, while Cutie achieves a $\mathcal{J\&F}$ score of 68.3\%. Despite Cutie's use of object memory to enhance object-level feature retention during segmentation and QDMN++'s emphasis on temporal consistency to address overlaps between similar objects, neither method demonstrates satisfactory results on this dataset. 

To advance the task of segmentation in dynamic video scenes, the 4th Pixel-level Video Understanding in the Wild (PVUW) Challenge \cite{ding2024pvuw} will be held in conjunction with CVPR 2025. This competition consists of two tracks: the MOSE \cite{ding2023mose} track aims to track and segment objects in videos captured in complex environments, while the MeViS \cite{ding2023mevis} track focuses on segmenting objects in videos based on textual references describing their motion. We chose to participate in the MOSE track as it aligned closely with the focus of our research. In the literature review, we found that the state-of-the-art open-source method, SAM 2, proposed by Ravi et al. \cite{ravi2024sam}, has been widely adopted. Therefore, we chose this method as the primary network framework for our solution, upon which we performed fine-tuning and further improvements. Specifically, we fine-tuned the SAM 2-Large model on MOSE. After identifying the optimal parameters on the validation set, we applied them to the test set. Furthermore, as SAM 2 segments individual objects and merges them afterward, this process may result in visible gaps at the boundaries of adjacent objects. To address this issue, we adopted morphological post-processing techniques, which alleviated the problem to a certain extent. Finally, we applied the optimal model to multi-scale images for predictions, and ensemble voting was used to fuse the resulting predictions to generate the final results. On the validation set, our best score reached 76.81\% $\mathcal{J\&F}$, while on the test set, our top score was 83.92\% $\mathcal{J\&F}$, ranking third on the leaderboard.

%% file: sec/2_formatting.tex
\section{Method}
\label{sec:formatting}

\begin{figure*}[!htbp]
	\centering
	\includegraphics[width=0.76\linewidth]{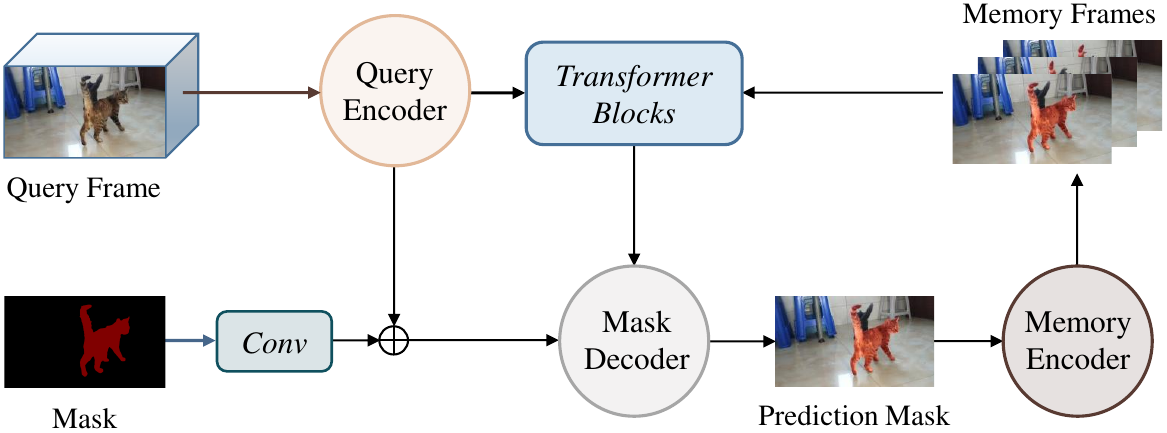}
	\caption{Network Architecture of our FVOS. It mainly consists of a query encoder, a memory encoder, a mask decoder and attention Transformer blocks.}
	\label{fig1}
\end{figure*}

\begin{figure*}[!htbp]
	\centering
	\includegraphics[width=0.92\linewidth]{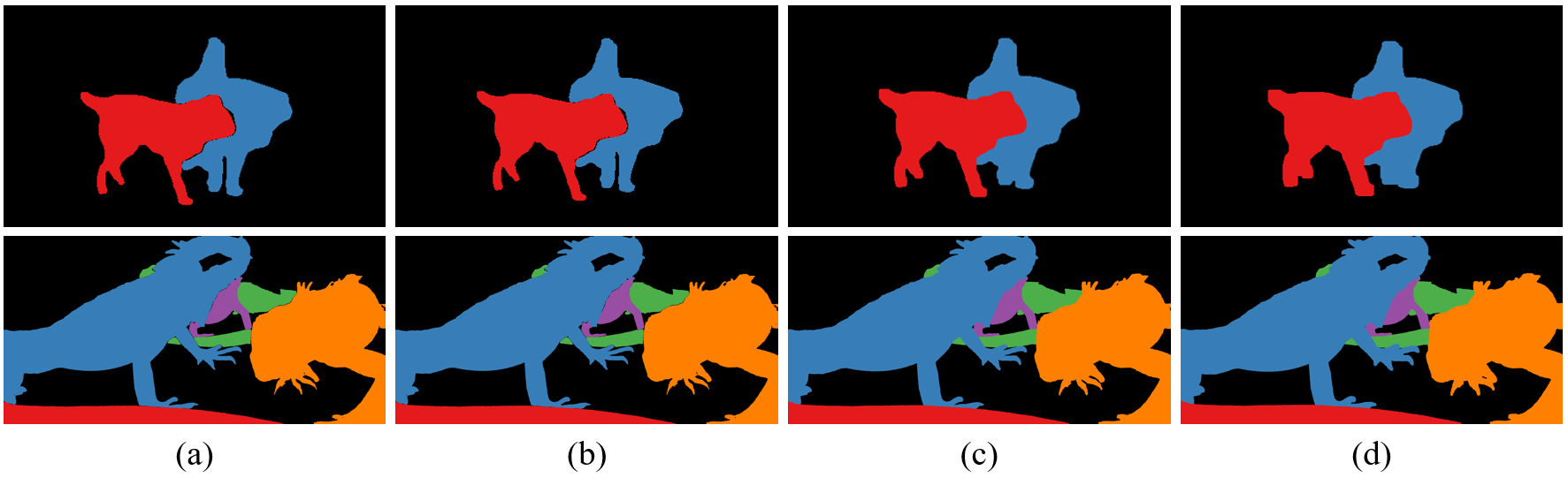}
	\caption{Morphological post-processing results on the MOSE test dataset. (a) kernel=0. (b) kernel=2. (c) kernel=3. (d) kernel=5.}
	\label{fig2}
\end{figure*}

Our approach primarily consists of three components: model fine-tuning training, morphological post-processing, and multi-scale segmentation result fusion. The model fine-tuning training process is detailed in Section \ref{2.1}, the morphological post-processing methodology is outlined in Section \ref{2.2}, and the fusion of multi-scale segmentation results is described in Section \ref{2.3}.

\subsection{MOSE Fine-tuning}
\label{2.1}

Figure \ref{fig1} illustrates the network architecture adopted in our framework, which primarily relies on Transformers for feature extraction and attention computation. Specifically, the current query frame is fed into the query encoder. The query encoder consists of a hierarchical Transformer pre-trained with a masked autoencoder. Next, the encoded query features are input into multiple stacked Transformer blocks for memory attention computation. On the other side, the input masked prompt undergoes convolutional embedding and is added to the frame embedding. The result of this addition, along with the memory attention computation outcome, is passed into the mask decoder. Similarly, the mask decoder comprises multiple stacked bidirectional Transformer blocks, designed to update the masked prompt and frame embedding. The resulting output mask is then fed into the memory encoder, which mainly consists of convolutional modules. Multiple frames processed by the memory encoder form a memory queue, which is stored as a memory bank and enhanced via skip connections for memory attention computation.

Our training process is as follows: First, we fine-tune the pre-trained model on the MOSE dataset for a total of 10 epochs, submitting results from the validation set of each epoch. The best-performing model from this stage is selected as the pre-trained model to begin a new round of training. In this second stage, we conduct training for a total of 40 epochs, selecting the best-performing model for testing with optimal parameters. Finally, the single best-performing model is selected to generate the initial single-model segmentation results.

\subsection{Morphological Post-Processing}
\label{2.2}

After performing the operations described in Section \ref{2.1}, we obtained the best results from a single model. Upon observing the results, we noticed that although the model provided satisfactory predictions in most cases, there is a notable issue: there exists a distinct gap between adjacent objects. This issue arises during the inference stage because the model predicts separate objects individually before merging them. As adjacent objects are treated independently, the edge regions are not well aligned. To address this problem, we propose using morphological operations for post-processing \cite{comer1999morphological}.

Morphological operations primarily extract structural information from within an image, which is often crucial for expressing and depicting the image's features. These features are typically the essential shape characteristics used in image understanding. Morphological methods have significant applications in areas such as visual detection, text recognition, medical image processing, and image compression and encoding. The key morphological operations include erosion, dilation, opening, closing, morphological gradient, top-hat transformation, and black-hat transformation. Among these, dilation and erosion serve as the fundamental morphological operations in image processing and are often combined to achieve more complex morphological manipulations.

We primarily adopt the method of dilation processing. During implementation, two parameters are generally required as inputs: one is the original image, and the other is referred to as a structural element or kernel, which determines the nature of the operation. Formalizing it, let ``$\oplus$'' denote dilation, and the definition is as follows:

\begin{equation}
A\oplus B = \left\{x|(B)_x\cap A\neq\Theta\right\}
\end{equation}

This formula represents the process of performing dilation on image $A$ using $B$, where $B$ is a convolution template or kernel. Its shape can be either square or circular. Through convolution computation between template $B$ and image $A$, each pixel in the image is scanned. A logical ``AND'' operation is performed between the template elements and the binary image elements. If all values are 0, the target pixel value is set to 0; otherwise, it is set to 1. In this way, the maximum pixel value in the region covered by $B$ is calculated, and the reference pixel value is replaced with this maximum value to achieve dilation.

During the inference process of the network, the binary Boolean segmentation masks for each object are first obtained and collected. For the current object, dilation operations are performed on both the object itself and all other objects. The adjacency between other objects and the current object is determined by checking whether the dilated masks overlap. If objects are deemed adjacent, the overlapping regions are filled and applied to the current object. Finally, object mask merging is performed following the rule of prioritizing higher-indexed objects, yielding the final segmentation results. 

Figure \ref{fig2} illustrates the results of applying different convolution kernel morphological post-processing techniques to the test set. It is evident that significant gaps exist between different objects in the initial results. However, as the kernel size increases, the connections between objects become more seamless. Nevertheless, the filling of gaps also introduces a certain degree of edge filling for the objects. Therefore, a larger kernel size is not always better. In fact, based on our experiments, using a kernel size of 2 yields better improvements in the segmentation results. 

\subsection{Multi-Scale Results Fusion}
\label{2.3}

\begin{figure}[!htbp]
	\centering
	\includegraphics[width=1\linewidth]{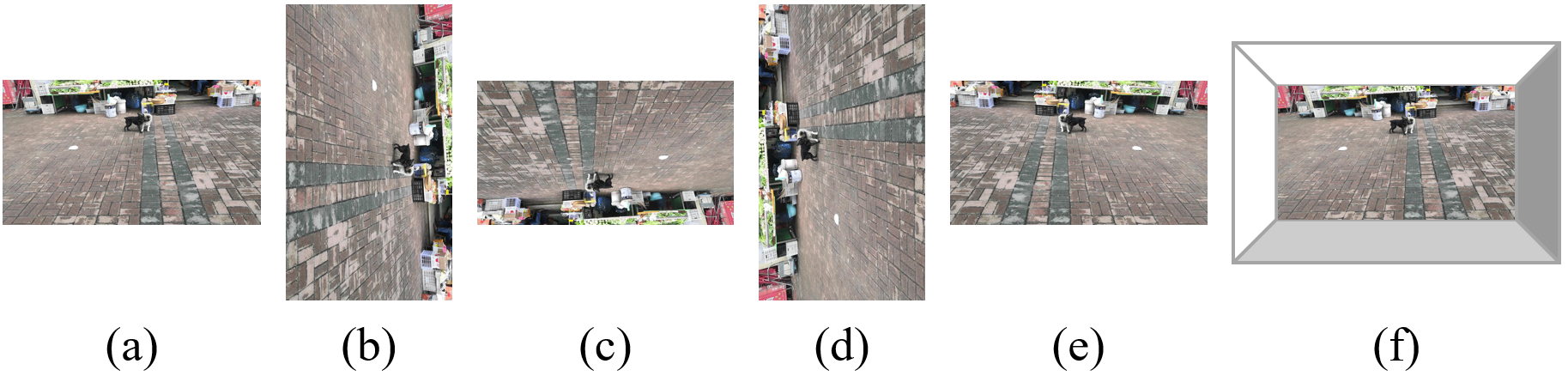}
	\caption{Test time data augmentation and multi-scale magnification operations. (a) original image. (b) clockwise by 90$^\circ$. (c) clockwise by 180$^\circ$. (d) clockwise by 270$^\circ$. (e) horizontal flipping. (f) multi-scale magnification.}
	\label{fig3}
\end{figure}

To further improve the results, we also adopted common test-time data augmentation methods, including rotating the original image clockwise by 90$^\circ$, 180$^\circ$, and 270$^\circ$, horizontal flipping, as well as multi-scale processing by resizing the image to several scales, as shown in Figure \ref{fig3}. Through experiments, we found that rotation and horizontal flipping did not enhance the results, possibly due to the lack of similarly processed data during training, which made the model less adaptable to such data changes during inference. Therefore, we focused primarily on the fusion of multi-scale results. 

Specifically, starting from the original size, we resized the dataset with increments of 0.125 to reconstruct it at multiple scales. After experimenting with several scales, we finally selected seven different scales ranging from 1 to 1.75 for fusion. 

%% file: sec/3_finalcopy.tex
\section{Experimental Results}
\label{sec3}

In this section, we apply the method elaborated in Section \ref{sec:formatting} to conduct experiments. First, we introduce the experimental data and evaluation metrics. Then, the experimental setup is introduced, including the experimental environment and implementation details. Finally, we present the final results comparison.

\subsection{Datasets}

{\textbf{\emph{MOSE}}} \cite{ding2023mose} primarily features highly challenging scenarios in a vast number of complex scenes. These include crowded target groups, various forms of occlusion, objects that disappear and reappear, and inconspicuous small objects. The dataset consists of 2,149 videos with a cumulative duration of 443 minutes, covering 36 categories and 5,200 objects. A total of 431,725 annotated masks are provided. Moreover, MOSE significantly outperforms other datasets in terms of annotation scale and total duration. 

\subsection{Evaluation Metrics}

In the VOS task, the two core evaluation metrics are the Jaccard Index ($\mathcal{J}$) and the $\mathcal{F}$-score. The Jaccard Index is used to measure the overlap between the predicted mask and the ground truth mask, while the F-score evaluates the consistency between the predicted boundary and the ground truth boundary. By balancing precision ($P$) and recall ($R$), the $\mathcal{F}$-score comprehensively reflects the segmentation accuracy.

\begin{equation}
F = \frac{2PR}{P+R} \ \ \  P = \frac{TP}{TP+FP} \ \ \ R = \frac{TP}{TP+FN}
\end{equation}

Among them, $TP$ represents true positives, $FP$ represents false positives, and $FN$ represents false negatives. 

\subsection{Experimental Environment}
All training experiments are experimentally performed on NVIDIA H800 (80GB) * 2 GPUs, with 40 vCPU Intel(R) Xeon(R) Platinum 8458P. The inference experiments are done on NVIDIA H800 (80GB) * 2 GPUs and NVIDIA RTX 3090 (24GB) * 1 GPU. The deep learning framework we use is PyTorch 2.1.0, the computing platform is Cuda 12.1, and the version of Python is Python 3.10 (Ubuntu 22.04).

\subsection{Implementation Details}

Our default parameter settings follow \cite{ravi2024sam}; therefore, we mainly describe here the optimal parameters we have identified and their meanings, as shown in Table \ref{otad}.

\begin{table}[htbp]
\centering
\caption{Optimal Parameter Settings for FVOS.}
\label{otad}
\begin{spacing}{1.2}
\begin{threeparttable}
\scalebox{0.62}{
\begin{tabular}{ccc}
\toprule[1.3pt]
Parameters & Meanings & Values\\
\hline
$epoch_v$ & Optimal Model in Validation & epoch-3 in second stage \\
$epoch_t$ & Optimal Model in Test & epoch-4 in second stage\\
$ssme_v$ & Sigmoid Scale for Memory Encoder in Validation & $21$\\
$ssme_t$ & Sigmoid Scale for Memory Encoder in Test & $20$\\
$sbme_v$ & Sigmoid Bias for Memory Encoder in Validation & $-10$\\
$sbme_t$ & Sigmoid Bias for Memory Encoder in Test & $-9$\\
\bottomrule[1.3pt]
\end{tabular}
}
\end{threeparttable}
\end{spacing}
\end{table}

\subsection{Performance Comparison}
Our method achieved 76.81\% $\mathcal{J\&F}$ on the validation set and 83.92\% $\mathcal{J\&F}$ on the test set, compared with the other methods on the leaderboard as Table \ref{table2} and Table \ref{table3}. It is important to note that since the leaderboard will overwrite the previous scores, the results we have written in the table are our best results, but there is no guarantee that the results we have seen for the other methods will also be optimal.

As can be seen from the tables, our method achieves good results on the validation set, but does not achieve very good results on the test set. We will continue to study this in the future to find further improvements.

\begin{table}[htbp]
\centering
\caption{Comparison of methods on the leaderboard in the validation phase.}
\label{table2}
\begin{spacing}{1.3}
\begin{threeparttable}
\scalebox{0.99}{
\begin{tabular}{cccc}
\toprule[1.35pt]
Methods & $\mathcal{J}$ & $\mathcal{F}$  & $\mathcal{J\&F}$  \\
\hline
\textbf{Ours} & \textbf{0.7278} & \textbf{0.8084} & \textbf{0.7681} \\
KirinCZW  &  0.7105 & 0.7878 & 0.7491   \\
xxxxl  &  0.7057 & 0.7845 & 0.7451  \\
civa\_lab  & 0.6865 & 0.7707 & 0.7286 \\
NanMu &  0.6579 & 0.7401 & 0.6990   \\
skh  &  0.6484 & 0.7289 & 0.6887 \\
ntuLC  & 0.5522 & 0.6375 & 0.5948 \\
\bottomrule[1.35pt]
\end{tabular}
}
\end{threeparttable}
\end{spacing}
\end{table}

\begin{table}[htbp]
\centering
\caption{Comparison of methods on the leaderboard in the test phase.}
\label{table3}
\begin{spacing}{1.3}
\begin{threeparttable}
\scalebox{0.99}{
\begin{tabular}{cccc}
\toprule[1.35pt]
Methods & $\mathcal{J}$ & $\mathcal{F}$  & $\mathcal{J\&F}$  \\
\hline
imaplus & 0.8359 & 0.9092 & 0.8726 \\
KirinCZW  &  0.8250 & 	0.9007 & 0.8628   \\
\textbf{Ours}  & \textbf{0.8028} & \textbf{0.8757} & \textbf{0.8392}  \\
SCU\_Leung  & 0.7993 & 0.8733 & 0.8363 \\
wulutuluman &  0.7989 & 0.8721 & 0.8355   \\
mima  &  0.7980 & 0.8721 & 0.8351 \\
LK18621786396  & 0.7980 & 0.8710 & 0.8345  \\
STELATOS9 & 0.7965 & 0.8716 & 0.8341 \\
\bottomrule[1.35pt]
\end{tabular}
}
\end{threeparttable}
\end{spacing}
\end{table}

%% file: sec/4_conclusion.tex
\section{Conclusion}
\label{sec4}

This paper primarily focuses on semi-supervised video object segmentation on complex datasets. We propose fine-tuning the state-of-the-art method specifically for MOSE dataset without utilizing any additional datasets. After obtaining the optimal single model, we apply morphological methods to post-process the prediction results. Furthermore, we experiment with various data augmentation techniques during the inference stage, ultimately selecting a multi-scale model fusion approach to produce the final results. Our FVOS achieves the first place during validation stage and the third place during testing stage in the MOSE Track of the 4th PVUW challenge 2025.